\documentclass[runningheads]{llncs}
\usepackage[T1]{fontenc}
\usepackage{graphicx}
\usepackage{marvosym}

\usepackage{amsmath,amssymb} 
\usepackage{color}
\usepackage{bm}
\usepackage{bbm}
\usepackage{booktabs}
\usepackage{multirow}
\usepackage{algorithm}
\usepackage[noend]{algorithmic} %
\usepackage{wrapfig}
\usepackage{hyperref}

\begin{document}
%
\title{Learning Inter-Superpoint Affinity for Weakly Supervised 3D Instance Segmentation}
\titlerunning{Weakly Supervised 3D Instance Segmentation}
%

\author{Linghua Tang \and
	Le Hui \and
	Jin Xie}
\authorrunning{L. Tang et al.}
%

\institute{Nanjing University of Science and Technology, Nanjing, China \\
	\email{ \{tanglinghua, le.hui, csjxie\}@njust.edu.cn }
	}

\maketitle              
\begin{abstract}
Due to the few annotated labels of 3D point clouds, how to learn discriminative features of point clouds to segment object instances is a challenging problem. In this paper, we propose a simple yet effective 3D instance segmentation framework that can achieve good performance by annotating only one point for each instance. Specifically, to tackle extremely few labels for instance segmentation, we first oversegment the point cloud into superpoints in an unsupervised manner and extend the point-level annotations to the superpoint level. Then, based on the superpoint graph, we propose an inter-superpoint affinity mining module that considers the semantic and spatial relations to adaptively learn inter-superpoint affinity to generate high-quality pseudo labels via semantic-aware random walk. Finally, we propose a volume-aware instance refinement module to segment high-quality instances by applying volume constraints of objects in clustering on the superpoint graph. Extensive experiments on the ScanNet-v2 and S3DIS datasets demonstrate that our method achieves state-of-the-art performance in the weakly supervised point cloud instance segmentation task, and even outperforms some fully supervised methods. Source code is available at \url{https://github.com/fpthink/3D-WSIS}.

\end{abstract}
\section{Introduction}
Point cloud instance segmentation is a classic task in 3D computer vision, and it can be applied in many fields, including indoor navigation systems, augmented reality, and robotics. The fully supervised instance segmentation methods~\cite{liang2021instance,chen2021hierarchical,jiang2020pointgroup,hui2022graphcut} have achieved impressive results, but they rely on numerous manually labeled data. However, annotating a large number of point clouds is extremely time-consuming and expensive. Thus, it is meaningful to segment point clouds in a semi-/weakly supervised manner that requires a small number of annotations. However, how to fully exploit the limited labels to improve the performance of instance segmentation is still a challenging problem.

Few efforts have been dedicated to semi-/weakly supervised point cloud instance segmentation. As a pioneer, Liao~\emph{et al.}~\cite{liao2021point} proposed a semi-supervised point cloud instance segmentation method using bounding boxes as supervision, where a network is used to generate bounding box proposals. And instance segmentation is achieved by refining the point cloud within the bounding box proposals. Besides, Tao~\emph{et al.}~\cite{tao2020seggroup} proposed a two-stage seg-level supervision 3D instance and semantic segmentation method, which first leverages a segment grouping network to generate pseudo labels for the whole scenes, and then the generated pseudo point-level labels are used as the ground truth to train the network. However, these simple pseudo label generation strategies cannot effectively generate high-quality pseudo labels, resulting in poor 3D instance segmentation results.

In this paper, we propose a simple yet effective weakly supervised 3D instance segmentation framework, which can achieve impressive results with one point annotation per instance. For weakly supervised point cloud instance segmentation with few annotated labels, our intuition lies in two folds: (1) Under rare annotations, effective label propagation is essential to produce high-quality pseudo labels, especially in 3D instance segmentation. (2) Weakly supervised 3D instance segmentation is more challenging than weakly supervised 3D semantic segmentation, so we consider introducing the object volume constraint to improve the instance segmentation results. Specifically, we first use an unsupervised method~\cite{landrieu2018large} to oversegment the point cloud into superpoints and build the superpoint graph. In this way, point-level labels can be extended to superpoint-level labels. Then, we propose an inter-superpoint affinity mining module to generate high-quality pseudo labels based on a few annotated superpoint-level labels. Based on the superpoint graph, we leverage the semantic and spatial information of adjacent superpoints to adaptively learn inter-superpoint affinity, which can be used to propagate superpoint labels along the superpoint graph via semantic-aware random walk. Finally, we propose a volume-aware instance refinement module to improve instance segmentation performance. Based on the trained model using superpoint-level propagation, we can obtain coarse instance segmentation results through superpoint clustering and further infer the object volume information from the instance segmentation results. The object volume information contains the number of voxels and the radius of the object. The inferred object volume information is regarded as the ground truth of the corresponding instance to retrain the network. In the test phase, based on the object volume information, we utilize the predicted object volume information to introduce a volume-aware instance clustering algorithm for segmenting high-quality instances. Extensive experiments on the ScanNet-v2~\cite{dai2017scannet} and S3DIS~\cite{armeni20163d} datasets can demonstrate the effectiveness of our method.


The main contributions of our paper are as follows: 
\begin{itemize}
	\item We present an inter-superpoint affinity mining module that considers the semantic and spatial relation to adaptively learn inter-superpoint affinity for random-walk based label propagation.
	\item We present a volume-aware instance refinement module, which guides the superpoint clustering on the superpoint graph to segment instances by using the object volume information.
	\item Our simple yet effective framework achieves state-of-the-art weakly supervised 3D instance segmentation performance on popular datasets ScanNet-v2 and S3DIS.
\end{itemize}

\section{Related Work}

\subsection{3D Semantic Segmentation}
{\bf Fully supervised 3D semantic segmentation}. Many methods have been proposed to achieve point cloud semantic segmentation. Some methods~\cite{lawin2017deep,tatarchenko2018tangent,jaritz2019multi} project point clouds into a series of regular 2D images from different views, and then fuse features extracted through 2D convolutional neural networks (CNNs). To apply 3D CNNs on the irregular point cloud and alleviate large memory costs, many efforts~\cite{graham20183d,choy20194d} first voxelize the point cloud into voxels and then utilize the sparse convolutional neural network to extract features of the point cloud. PointNet~\cite{qi2017pointnet} directly extracts features from points with shared multi-layer perceptrons and max-pooling layer. Inspired by PointNet, different local feature aggregation operators~\cite{qi2017pointnet++,thomas2019kpconv,wu2019pointconv,cheng2020cascaded} are proposed to work on point cloud, which directly consume point cloud. Besides, various methods~\cite{wang2019dynamic,hui2021superpoint} capture intrinsic spatial and geometric features by constructing the graph on the point cloud. Various approaches exploit different local feature aggregation networks to extract discriminative point features and use multi-layer perceptrons to achieve 3D semantic segmentation.



{\bf Semi-/Weakly supervised 3D semantic segmentation}. Inspired by class activation map in 2D images, Wei~\emph{et al.}~\cite{wei2020multi} introduce a multi-path region mining module to generate pseudo labels, which only requires cloud-level weak labels. Xu~\emph{et al.}~\cite{xu2020weakly} use three additional losses to constrain on unlabeled points, achieving impressive performance with 10\% labels. Cheng~\emph{et al.}~\cite{cheng2021sspc} use a dynamic label propagation strategy to generate pseudo labels, and learn discriminative features with a coupled attention module. Zhang~\emph{et al.}~\cite{zhang2021perturbed} exploit the consistency generated by perturbation to obtain additional supervision and propagate implicit labels by constructing the graph topology of the point cloud. Liu~\emph{et al.}~\cite{liu2021one} first build a supervoxel graph on the point cloud and then conduct label propagation by learning the similarity among graph nodes. Li~\emph{et al.}~\cite{li2022hybridcr} utilize a hybrid contrastive regularization strategy with point cloud augmentation to provide additional constraints for network training. To generate pseudo labels for outdoor point cloud scenes,  Shi~\emph{et al.}~\cite{shi2022weakly} design a matching module to propagate pseudo labels in both temporal and spatial spaces.

\subsection{3D Instance Segmentation}
{\bf Fully supervised 3D instance segmentation}. Compared with point cloud semantic segmentation, instance segmentation is more challenging because it not only requires predicting semantic scores but also distinguishing instances of the same class. According to the different manners of generating instances, instance segmentation methods can be mainly divided into clustering-based methods and proposal-based methods. Given point clouds as input, clustering-based methods regard instance segmentation as the post-processing task after network inference, and the result is obtained by clustering on point clouds with the predicted features. As a pioneer, Wang~\emph{et al.}~\cite{wang2018sgpn} introduce a similarity matrix to measure the distances between the features of all point pairs, which guides clustering points as proposals. Wang~\emph{et al.}~\cite{wang2019associatively} integrate semantic and instance segmentation into a parallel framework, which benefits from each other task. Lahoud~\emph{et al.}~\cite{lahoud20193d} design a multi-task neural network architecture, where instances are simultaneously separated in the feature vector space and direction vector space by a discriminative loss~\cite{de2017semantic} and a directional loss. Jiang~\emph{et al.}~\cite{jiang2020pointgroup} generate proposals by clustering points on the original and offset-shifted coordinate spaces, which benefits from both advantages. Hou~\emph{et al.}~\cite{hou20193d} jointly learn color and geometry features for instance segmentation from different modalities. Lately, Chen~\emph{et al.}~\cite{chen2021hierarchical} introduce a hierarchical aggregation method that iteratively clusters point clouds into instance proposals. Liang~\emph{et al.}~\cite{liang2021instance} propose a semantic superpoint tree structure and achieved instance segmentation by tree traversal and splitting. Vu~\emph{et al.}~\cite{vu2022softgroup} design a soft group algorithm to reduce the semantic prediction errors to significantly boost the segmentation performance.

For proposal-based methods, instance segmentation consists of two procedures, first generating rough proposals and then predicting precise instance masks. Yang~\emph{et al.}~\cite{yang2019learning} propose an end-to-end trainable network which directly generates 3D bounding boxes as proposals and infers point-wise instance masks for points inside proposals. Instead of getting proposals via 3D bounding box regression, Yi~\emph{et al.}~\cite{yi2019gspn} introduce an approach to obtain proposals by object generation, and then predict instance masks within proposals.

{\bf Semi-/Weakly supervised 3D instance segmentation}. Few efforts have been made on semi-/weakly supervised point cloud instance segmentation. Tao~\emph{et al.}~\cite{tao2020seggroup} propose a method to generate pseudo labels for the whole training scene and the generated pseudo point-level labels are used to train existing full supervised methods for point cloud instance segmentation, where one point per instance is clicked as the weak label. Nonetheless, the quality of pseudo labels is limited due to lack of learning discriminative instance features. With bounding boxes as weak labels, Liao~\emph{et al.}~\cite{liao2021point} propose a semi-supervised point cloud instance segmentation method, where a network is leveraged to generate bounding box proposals and instance segmentation is achieved by refining points within bounding box proposals.


\section{Method}
\begin{figure*}[t]
	\centering
	\includegraphics[width=1\textwidth]{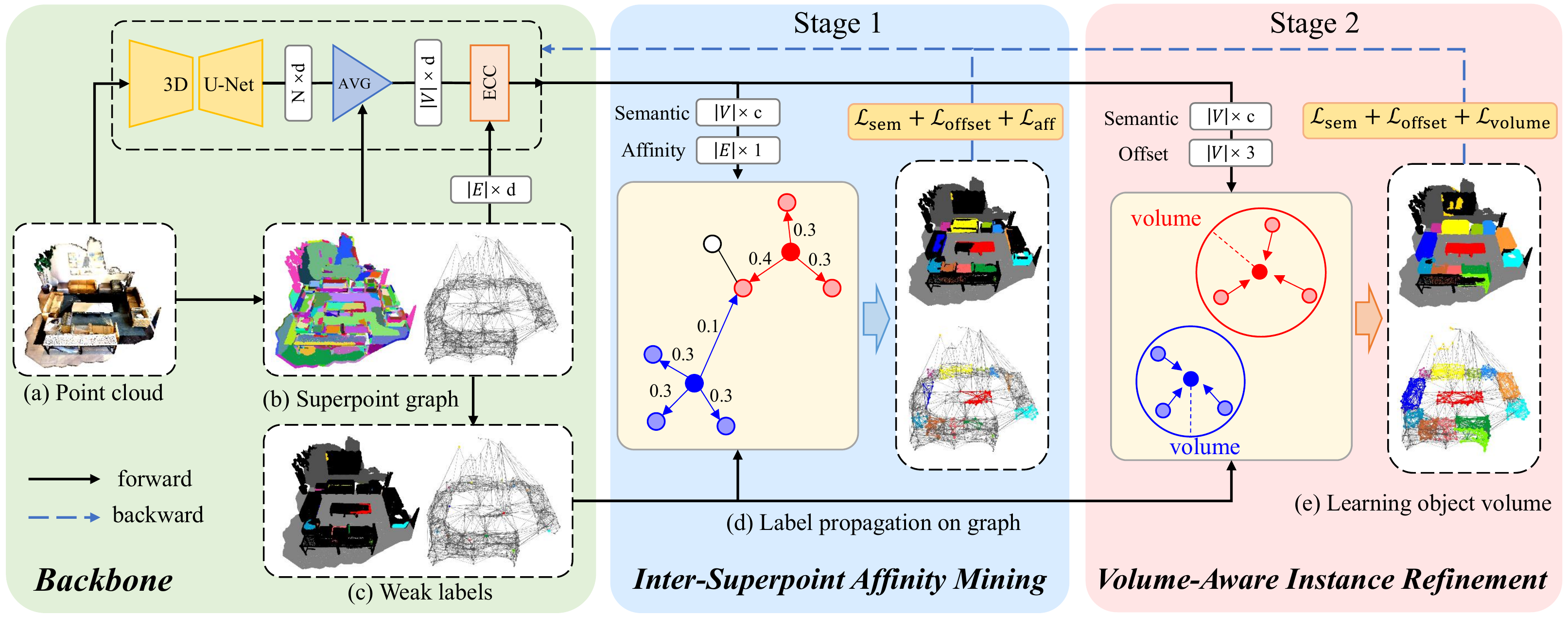}
	\caption{Overview of our framework for weakly supervised 3D instance segmentation. We first oversegment point cloud (a) to build the superpoint graph (b) and extend weak labels (c) to the corresponding superpoint. Then, based on superpoint graph, the random walk with the predicted affinity and semantic is used for label propagation (d). Finally, combining the predicted semantic and offset, learning pseudo object volume (e) is achieved. $c$ is the number of categories, $d$ is the feature dimension, $N$ is the number of points, $|V|$ is the number of superpoints, and $|E|$ is the number of edges.}
	\label{fig:pipeline}
\end{figure*}

The overall architecture of our method is depicted in Fig.~\ref{fig:pipeline}. The backbone network (Sec.~\ref{sec:backbone}) first takes the point cloud and superpoint graph as input and predicts superpoint-wise semantic labels and offset vectors. Then, the inter-superpoint affinity mining module (Sec.~\ref{sec:affinity}) propagates labels on the superpoint graph via semantic-aware random walk. Finally, the volume-aware instance refinement module (Sec.~\ref{sec:ins_size}) learns object volume information to improve instance segmentation performance.

\subsection{Backbone Network}\label{sec:backbone}
{\bf Superpoint graph construction.} Following~\cite{landrieu2018large,liang2021instance}, we adopt an unsupervised point cloud oversegmentation method to generate the superpoints and construct the superpoint graph. The superpoint graph is a geometry representation of point clouds, defined as $G=\left(V, E\right)$. Vertice $V$ means superpoints that are generated by aggregating points with similar geometric characteristics, and edge $E$ indicates the prior connection relationship between adjacency superpoints which are constructed by linking $k$-nearest superpoints. In weakly supervised 3D instance segmentation, the benefits of the use of superpoints are two-fold. On one hand, the superpoint is a geometrically homogeneous unit, so we can extend annotated point label to the corresponding superpoint, thereby alleviating the sparsity of point-level annotations. On the other hand, the superpoint graph captures the spatial relationship of different instances so that we can utilize it to perform label propagation efficiently.

{\bf Superpoint feature extraction.} Specifically, we first extract point features using 3D U-Net~\cite{graham20183d} on the point cloud and then aggregate point features into superpoint features by average pooling. After that, based on the superpoint graph, we use the edge-conditioned convolutions (ECC)~\cite{simonovsky2017dynamic} to extract superpoint features. Finally, we use the superpoint features to predict the semantic label of superpoints.

\subsection{Inter-Superpoint Affinity Mining}\label{sec:affinity}
To perform label propagation, we develop an inter-superpoint affinity mining module to learn the superpoint relationship in the semantic and coordinate spaces. By using the learned inter-superpoint affinity, we design a simple semantic-aware random walk algorithm for label propagation on the superpoint graph.

{\bf Superpoint affinity learning.} Based on the superpoint graph, we learn the relationship between two adjacent superpoints to characterize their affinity. It is desired that the learned affinity between two adjacent superpoints can guide the label propagation along the edge of the superpoint graph. Assuming the learned superpoint embedding in the backbone network is $\bm{X}\in\mathbb{R}^{|V|\times d}$, where $|V|$ is the number of superpoints and $d$ is the feature dimension. Given the $i$-th superpoint embedding $\bm{X}_{i}\in\mathbb{R}^{d}$ and its first-order neighbors $\mathcal{N}_i$, we leverage the semantic and spatial information of the superpoints to adaptively learn inter-superpoint affinity. The affinity $A_{ij}$ between the $i$-th superpoint and its $j$-th neighbor is formulated as:
\begin{equation}
	\setlength{\abovedisplayskip}{1pt}
	\setlength{\belowdisplayskip}{1pt}
	A_{ij} = \frac{\exp(\sigma(\phi(\bm{X}_{i}), \psi( \bm{X}_{j})) * \gamma(\bm{p}_i-\bm{p}_j))}{ \sum_{k \in \mathcal{N}_i} \exp(\sigma(\phi(\bm{X}_{i}), \psi(\bm{X}_{k})) *  \gamma(\bm{p}_i-\bm{p}_k))}
	\label{eqn:affinity}
\end{equation}
where $\bm{X}_{i}\in\mathbb{R}^{d}$ and $\bm{X}_{j}\in\mathbb{R}^{d}$ are the superpoint embedding. $\bm{p}_i\in\mathbb{R}^3$ and $\bm{p}_j\in\mathbb{R}^3$ are the centroid coordinate of the superpoints. $\phi(\cdot)$ and $\psi(\cdot)$ are linear projections, and $\gamma(\cdot)$ is a multi-layer perceptron. $\sigma(\cdot,\cdot)$ is the dot production for learning the similarity of the $i$-th and $g$-th superpoints. In~Eq.~(\ref{eqn:affinity}), semantic similarity is measured by dot production while spatial similarity is measured by subtraction. As a result, the affinity $A_{ij}$ considers the semantic and spatial information of the superpoints. After that, we use the learned inter-superpoint affinity to update the superpoint embeddings. For the $i$-th superpoint, the new superpoint embedding $\widetilde{\bm{X}}_{i}\in\mathbb{R}^{d}$ is written as:
\begin{equation}
	\setlength{\abovedisplayskip}{1pt}
	\setlength{\belowdisplayskip}{1pt}
	\widetilde{\bm{X}}_{i} = A_{ij} \cdot \rho (\bm{X}_{j}) + \bm{X}_{i}
	\label{eqn:new_sp_emb}
\end{equation}
where $\rho(\cdot)$ is linear projection. During the training, we employ a discriminative loss (dubbed $\mathcal{L}_{\text{aff}}$) used in~\cite{de2017semantic} to draw $\widetilde{\bm{X}}$ belonging to the same object towards each other, and make $\widetilde{\bm{X}}$ in different objects away. It is expected that the affinity $A_{ij}$ between the superpoints of the same instance can be enhanced.

\begin{figure*}[t]
	\centering
	\includegraphics[width=0.9\textwidth]{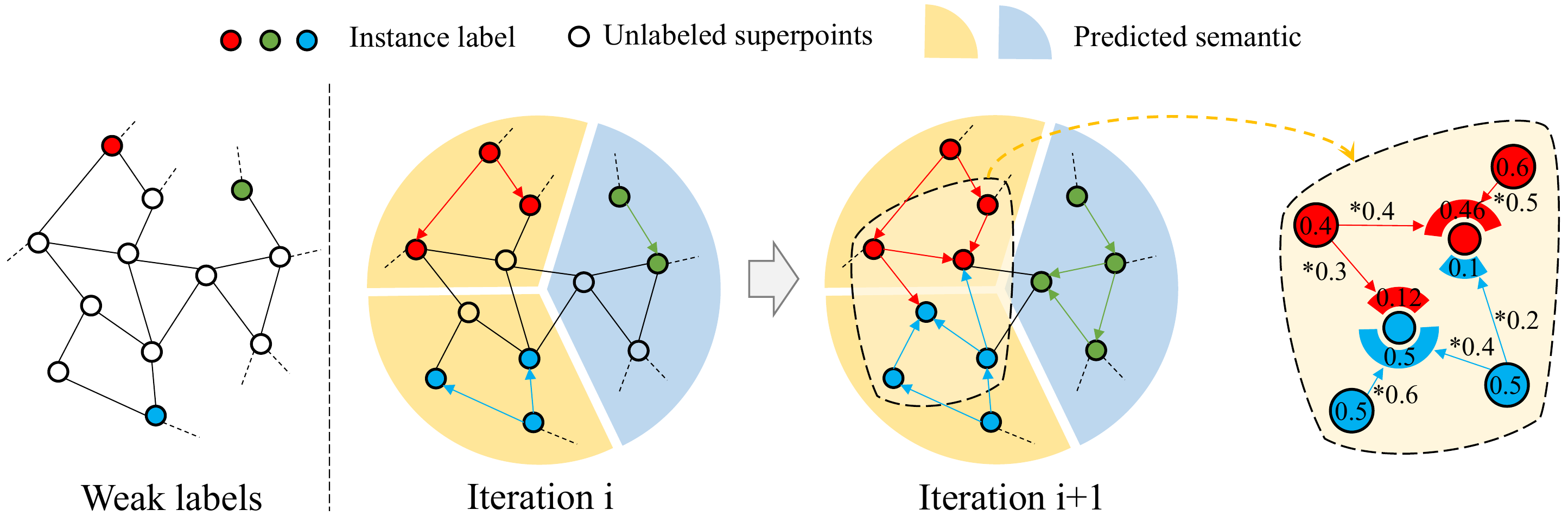}
	\caption{The process of label propagation with predicted instance affinity and semantics.}
	\label{fig:label_prop_declare}
\end{figure*}

{\bf Label propagation via semantic-aware random walk.} After obtaining the inter-superpoint affinity $\bm{A}\in\mathbb{R}^{|V|\times|V|}$, we design a simple semantic-aware random walk to propagate labels over the superpoint graph, as shown in Fig.~\ref{fig:label_prop_declare}. Specifically, our semantic-aware random walk propagates labels over the superpoints graph with the same predicted semantic labels. For the $c$-th class, assume that its semantic matrix is $\bm{S}^c\in\mathbb{R}^{|V|\times|V|}$. In $\bm{S}^c$, if the semantic class of the $i$-th and $j$-th superpoints are the same, then $S^{c}_{ij}=1$, otherwise $ S^{c}_{ij}=0$. For label propagation, we first using the semantic matrix $\bm{S}^c$, superpoint affinity $\bm{A}\in\mathbb{R}^{|V|\times|V|}$ and adjacency matrix $\bm{M}\in\mathbb{R}^{|V|\times|V|}$ of the graph $G=(V,E)$ to compute the weight $\bm{P}^c\in\mathbb{R}^{|V|\times|V|}$ for the $c$-th class, which is formulated as:
\begin{equation}
	\setlength{\abovedisplayskip}{1pt}
	\setlength{\belowdisplayskip}{1pt}
	\bm{P}^c = \bm{M} \odot \bm{S}^{c} \odot \bm{A}
	\label{eqn:mat_p}
\end{equation}
where $\odot$ is Hadamard product. Note that the weight $\bm{P}^c$ considers the semantic information and superpoint affinity simultaneously. Then, we derive the transition probability matrix $\bm{T}^c\in\mathbb{R}^{|V|\times|V|}$ for the $c$-th class, and it is defined as:
\begin{equation}
	\setlength{\abovedisplayskip}{1pt}
	\setlength{\belowdisplayskip}{1pt}
	\bm{T}^c = \bm{D}^{-1}\bm{P}^c, \ \text{where} \ D_{ii} = \sum\nolimits_{j} P_{ij}^c
\end{equation}
The diagonal matrix $\bm{D}$ is used for the row normalization of the matrix $\bm{P}^c$. Finally, the pseudo instance label $I_j$ of the $j$-th superpoint is propagated by:
\begin{equation}
	\setlength{\abovedisplayskip}{1pt}
	\setlength{\belowdisplayskip}{1pt}
	I_j =I_k, \text{where}~ k=\mathop{\operatorname{argmax}}\limits_{i=1,\ldots,|V|}( \bm{\hat{T}}^c_{ij} ) , \ \bm{\hat{T}}^c = (\bm{T}^c)^{t}
	\label{eqn:pl}
\end{equation}
where $\bm{\hat{T}}^c_{ij}$ indicates the probability of propagating the instance label of the $i$-th superpoint to the $j$-th superpoint, and $t$ is the iteration number. In Eq.~(\ref{eqn:pl}), the $j$-th superpoint selects the instance label of the $k$-th superpoint, which has the highest probability of propagating to the $j$-th superpoint. In this way, we can propagate the instance label of each annotated superpoint to unlabeled superpoints on the superpoint graph.

\subsection{Volume-Aware Instance Refinement}\label{sec:ins_size}
We propose the volume-aware instance refinement module to segment instances by using object volume information. We first introduce how to predict object volume via pseudo instances. Then, we present the volume-aware instance clustering algorithm to generate instances on the superpoint graph.

{\bf Object volume prediction via pseudo instance.} In order to predict object volume information, we first train the network with the pseudo labels generated by label propagation in the first stage. As shown in Fig.~\ref{fig:pipeline}~(e), we use the pre-trained model in the first stage to generate the pseudo instances by voting the superpoints to the closest annotated point. Specifically, by adding the predicted offset vector (refer to the first stage in Sec.~\ref{sec:network_training}) to the corresponding superpoint center, we can shift each superpoint center closer to the center of the corresponding object. To generate instances, the shifted superpoints are assigned the same instance labels as the closest annotated points with the same semantic labels. Here we regard the generated instances as the pseudo instances. According to the generated pseudo instances, we compute its volume information. We consider the number of voxels inside the instance and the instance radius to measure the volume of the instance. The instance radius is defined as the distance between the instance center and the farthest point. Thus, for each pseudo instance, we can obtain its volume information after the first stage.


\begin{figure}[!t]
	\centering
	\begin{minipage}[t]{.85\linewidth}
		
		\begin{algorithm}[H]
			\caption{Volume-Aware Instance Clustering Algorithm.}
			
			\textbf{Input:}~superpoint shifted coordinate $\{ \widetilde{\bm{p}}_1, \ldots, \widetilde{\bm{p}}_{|V|} \} \in \mathbb{R}^{|V| \times 3} $; ~superpoint semantic label $\{ s_1, \ldots, s_{|V|} \} \in \mathbb{R}^{|V| \times C}$;~the predicted voxel number $\{ u_1, \ldots, u_{|V|} \} \in \mathbb{R}^{|V|}$;~the predicted radius $\{ r_1, \ldots, r_{|V|} \} \in \mathbb{R}^{|V|}$
			
			\textbf{Output:}~generated instances $\textbf{I} \in \{ I_1, \ldots, I_m \}$,  $m$ is the number of instances.

			\begin{algorithmic}[1] \label{algo:bfs_cluster}
				\STATE Initialize an array $f$ (visited flag) of length $|V|$ with all zeros
				\STATE Initialize an empty proposal set $H$, an empty instance set \textbf{I}
				\STATE // Using the predicted radius $r$ to filter superpoints
				\FOR{ $v=1$ to $|V|$ }
				\IF{ $f_v == 0$ }
				\STATE Initialize an empty queue $Q$
				\STATE Initialize a set $H$
				\STATE $f_v = 1$ ; $Q$.enqueue($v$) ; add $v$ to $H$
				\WHILE{ $Q$ is not empty }
				\STATE $j=Q$.dequeue() 
				
				\FOR{ each $k \in \{ k \mid k \in \mathcal{N}_j, s_k == s_j, \Vert\widetilde{\bm{p}}_k - \widetilde{\bm{p}}_j \Vert_{2} < \lambda r_j \}$ }
				\IF{ $f_k == 0$ }
				\STATE $f_k = 1$ ; $Q$.enqueue($k$) ; add $k$ to $H$
				\ENDIF
				\ENDFOR
				\ENDWHILE
				
				\STATE add $H$ to \textbf{H}
				\ENDIF
				\ENDFOR
				
				\STATE // Using the predicted voxel numbers $u$ to filter proposals 
				\FOR{ each $ H \in \textbf{H} $}
				\STATE compute $\bar{w}$ = avg($\{u_i \mid i \in H \}$), $w$ for $H$
				\IF{ $w > \beta \bar{w}$}
				\STATE add $H$ to \textbf{I}
				\ENDIF
				
				\ENDFOR
				
				\FOR{ each $ H \in \textbf{H} $}
				\STATE compute $\bar{w}$ = avg($\{u_i \mid i \in H \}$), $w$ for $H$
				\IF{ $w \leq \beta \bar{w}$}
				\STATE $ I_{closest} $ = findClosestInstance($\{ I \mid I \in \textbf{I}, s_I == s_H \}$)
				\STATE $ I_{closest} = I_{closest} \cup H $
				\ENDIF
				
				\ENDFOR
				
				\RETURN \textbf{I}
			\end{algorithmic}
			
		\end{algorithm}
	\end{minipage}
\end{figure}

{\bf Volume-aware instance clustering.} After predicting the volume of the object in the first stage, we additionally use the predicted volume as the supervision to retrain the network (refer to the second stage in Sec.~\ref{sec:network_training}), which is regarded as the second stage. Thus, in the second stage, we can additionally predict the instance volume information (the number of voxels and the radius) for each superpoint. For the $i$-th superpoint, assume that the predicted semantic is $s_i\in\mathbb{R}^{1\times C}$, offset vector is $o_i\in\mathbb{R}^3$, the number of voxels is $u_i$, and the radius is $r_i$. Note that the predicted semantic $s_i$ is the one hot label. We first obtain the shifted coordinate of the superpoint by $\widetilde{\bm{p}}_i = \bm{p}_i + \bm{o}_i$, which makes the superpoint close to the corresponding instance center. Based on the shifted coordinate and the graph structure, the $i$-th superpoint merges its neighbors $\left\{ j \mid j \in \mathcal{N}_i, s_i = s_j, \Vert\widetilde{\bm{p}}_i - \widetilde{\bm{p}}_j \Vert_{2} < \lambda r_i \right\}$ into the same cluster, where the hyperparameter $\lambda$ is set to 0.25 empirically. Note that the radius $r_i$ is used to filter the superpoints far from the object center. Here, we use the breadth-first search on the superpoint graph to group nodes in the same cluster for generating compact instance proposals. After that, we further count the number of voxels $w$ in the proposal to filter the fragmented proposals. The predicted number of voxels $\bar{w}$ of the proposal is computed by averaging the predicted voxel numbers of superpoints within the proposal. If $w > \beta \bar{w}$, the corresponding proposal can be regarded as the instance. The hyperparameter $\beta$ is empirically set to 0.3. Finally, the remaining proposals are aggregated to the closest instance with the same semantic label. The volume-aware instance clustering algorithm is shown in Algorithm~\ref{algo:bfs_cluster}.

\subsection{Network Training}\label{sec:network_training}
Our method is a two-stage framework. As shown in Fig.~\ref{fig:pipeline}, the first stage learns the inter-superpoint affinity to propagate labels via random walk, while the second stage leverage the object volume information to refine the instance.

\textbf{First stage.} As shown in Fig.~\ref{fig:pipeline}, the first stage is supervised by the semantic loss $\mathcal{L}_{\text{sem}}$, offset loss $\mathcal{L}_{\text{offset}}$, and affinity loss $\mathcal{L}_{\text{aff}}$. The semantic loss $\mathcal{L}_{\text{sem}}$ is defined as the cross-entropy loss:
\begin{equation}
	\setlength{\abovedisplayskip}{1pt}
	\setlength{\belowdisplayskip}{1pt}
	\mathcal{L}_{\text{sem}} =  \frac{1}{ \sum\nolimits_{i=1}^{|V|} \mathbbm{I}(v_i) } \sum\nolimits_{i=1}^{|V|} \text{CE}(s_i, s_i^*) \cdot \mathbbm{I}(v_i)
\end{equation}
where $s_i$ is the predicted label and $s_i^*$ is the ground truth label. Note that the original annotated labels and generated pseudo labels are all regarded as the ground truth labels. If the superpoint $v_i$ has the label or is assigned with the pseudo label, the indicator function $\mathbbm{I}(v_i)$ is equal to 1, otherwise 0. In addition, we use MLP to predict the offset vector $\bm{o}_i \in \mathbb{R}^{3}$. the offset loss $\mathcal{L}_{\text{offset}}$ is used to minimize the predicted offset of the superpoint to its instance center. $\mathcal{L}_{\text{offset}}$ is defined as:
\begin{equation}
	\setlength{\abovedisplayskip}{1pt}
	\setlength{\belowdisplayskip}{1pt}
	\mathcal{L}_{\text{offset}} =  \frac{1}{ \sum\nolimits_{i=1}^{|V|} \mathbbm{I}(v_i) } \sum\nolimits_{i=1}^{|V|} {\Vert \bm{o}_i - \bm{o}_i^* \Vert}_1 \cdot \mathbbm{I}(v_i)
\end{equation}
where the $\bm{o}_i$ is the predicted superpoint offset and the $\bm{o}_i^*$ is the ground truth offset. Note the $\bm{o}_i^*$ is computed by coarse pseudo instance labels. Following~\cite{de2017semantic}, the affinity loss $\mathcal{L}_{\text{aff}}$ (refer to Sec.~\ref{sec:affinity}) is formulated as:

\begin{equation}
	\setlength{\abovedisplayskip}{0pt}
	\setlength{\belowdisplayskip}{0pt}
	\begin{aligned}
		\mathcal{L}_{\text{var}} = \frac{1}{I} \sum_{i=1}^{I} \frac{1}{ \sum_{j=1}^{|V|} \mathbbm{I}(v_j, i) } \sum\nolimits_{j=1}^{|V|} {\left[ {\Vert {\bm{\mu}}_i - \widetilde{\bm{X}}_{j} \Vert}_2 - {\delta}_v \right]}_+^2 \cdot \mathbbm{I}(v_j, i)\\
		\mathcal{L}_{\text{dist}} = \frac{1}{I(I-1)}   \mathop{ \sum\nolimits_{i_A=1}^{I} \sum\nolimits_{i_B=1}^{I} }\limits_{i_A \neq i_B} {\left[  2{\delta}_d - {\Vert {\bm{\mu}}_{i_A} - {\bm{\mu}}_{i_B} \Vert}_2 \right]}_+^2 \\
		\mathcal{L}_{\text{aff}} = \mathcal{L}_{\text{var}} + \mathcal{L}_{\text{dist}} + \alpha\cdot\mathcal{L}_{\text{reg}}, \text{where}~\mathcal{L}_{\text{reg}} = \frac{1}{I} \sum\nolimits_{i=1}^{I} {\Vert {\bm{\mu}}_i \Vert}_2
	\end{aligned}
\end{equation}
where $I$ is the number of instances (equal to the number of the annotated points, \emph{i.e}, one point per instance). ${\bm{\mu}}_i$ is the mean embedding of the $i$-th instance and $\widetilde{\bm{X}}_{j}$ is the embedding of the $j$-th superpoint in Eq.~(\ref{eqn:new_sp_emb}). According to~\cite{de2017semantic}, the margins ${\delta}_v$ and ${\delta}_d$ are set to 0.1 and 1.5, respectively. The parameter $\alpha$ is set to 0.001, and ${\left[x\right]}_+ = max(0, x)$ denotes the hinge. $\mathbbm{I}(v_j, i)$ is the indicator function, and $\mathbbm{I}(v_j, i)$ equals to 1 if superpoint $v_j$ is labeled as the $i$-th instance. Note that we only perform $\mathcal{L}_{\text{aff}}$ on the superpoints with annotated labels or pseudo labels. The final loss function in the first stage is defined as:
\begin{equation}
	\setlength{\abovedisplayskip}{1pt}
	\setlength{\belowdisplayskip}{1pt}
	\mathcal{L}_{\text{stage1}} = \mathcal{L}_{\text{sem}} + \mathcal{L}_{\text{offset}} + \mathcal{L}_{\text{aff}}
\end{equation}

\textbf{Second stage.} As shown in Fig.~\ref{fig:pipeline}, the second stage is supervised by the semantic loss $\mathcal{L}_{\text{sem}}$, offset loss $\mathcal{L}_{\text{offset}}$, and volume loss $\mathcal{L}_{\text{volume}}$. As the affinity loss is used for label propagation, we remove it in the second stage. For the volume loss $\mathcal{L}_{\text{volume}}$, it uses the predicted object volume information as the ground truth to train the network (refer to Sec.~\ref{sec:ins_size}). The $\mathcal{L}_{\text{volume}}$ is formulated as:
\begin{equation}
	\setlength{\abovedisplayskip}{1pt}
	\setlength{\belowdisplayskip}{1pt}
	\mathcal{L}_{\text{volume}} = \frac{1}{K} \sum\nolimits_{i=1}^{K}\sum\nolimits_{j=1}^{I}( {\Vert u_i - \hat{u}_j \Vert}_1 + {\Vert r_i - \hat{r}_j \Vert}_1 )\cdot \mathbbm{I}(i,j)
	\label{eq:size}
\end{equation}
where $K$ is the number of labeled superpoints, including the original annotated labels and the generated pseudo labels. If the $i$-th superpoint belongs to the $j$-th instance, the indicator function $\mathbbm{I}(i,j)$ is equal to 1, otherwise 0. $\hat{u}_j$ and $\hat{r}_j$ indicate the ground truth voxel numbers and radius counted from the pseudo instances, respectively. The generation of the pseudo instances refers to Sec.~\ref{sec:ins_size}. The final loss function in the second stage is defined as:
\begin{equation}
	\setlength{\abovedisplayskip}{1pt}
	\setlength{\belowdisplayskip}{1pt}
	\mathcal{L}_{\text{stage2}} = \mathcal{L}_{\text{sem}} + \mathcal{L}_{\text{offset}} + \mathcal{L}_{\text{volume}}
\end{equation}

\section{Experiments}

\subsection{Experimental Settings}
{\bf Datasets.} ScanNet-v2~\cite{dai2017scannet} and S3DIS~\cite{armeni20163d} are used in our experiments to conduct 3D instance segmentation. ScanNet-v2 contains 1,613 indoor RGB-D scans with dense semantic and instance annotations. The dataset is split into 1,201 training scenes, 312 validation scenes, and 100 hidden test scenes. The instance segmentation is evaluated on 18 object categories. S3DIS contains 6 large-scale indoor areas, which has 272 rooms and 13 categories. For the ScanNet-v2 dataset, we report both validation and online test results. For the S3DIS dataset, we report both Area 5 and the 6-fold cross validation results.


{\bf Evaluation metrics.} For the ScanNet-v2 dataset, the mean average precision at the overlap 0.25 ($\text{AP}_{25}$), 0.5 ($\text{AP}_{50}$) and overlaps from 0.5 to 0.95 (AP) are reported. For the S3DIS dataset, we additionally use mean coverage (mCov), mean weighted coverage (mWCov), mean precision (mPrec), and mean recall (mRec) with the IoU threshold of 0.5 as evaluation metrics.

{\bf Annotation of weak labels.} To generate weak labels of point clouds, we randomly click one point of each instance as the ground truth label. Note that our annotation strategy is the same as SegGroup~\cite{tao2020seggroup}. Unlike our method and SegGroup, SPIB~\cite{liao2021point} adopts 3D box-level annotation, which annotates each instance with bounding box. Compared with time-consuming box-level annotation, clicking one point per instance is faster and more convenient. 


\begin{table}[!t]
	\centering
	\caption{3D instance segmentation results on the ScanNet-v2 validation set and online test set. ``Baseline'' means the model trained with the initial annotated labels only.}
	\resizebox{0.95\textwidth}{!}{
		\begin{tabular}{@{}ccc|ccc|ccc@{}}
			\toprule
			\multicolumn{2}{c|}{Method}                                                    & \;Annotation\; & \;\;AP\;\;       & \;$\text{AP}_{50}$\;      & \;$\text{AP}_{25}$\;\;      & \;\;AP\;\;       & \;$\text{AP}_{50}$\;      & \;$\text{AP}_{25}$\;\;      \\
			\midrule
			\multicolumn{2}{c}{}                                                           &            & \multicolumn{3}{c|}{Validation Set} & \multicolumn{3}{c}{Online Test Set} \\ \midrule
			\multicolumn{1}{c|}{\multirow{7}{*}{Fully Sup.}}                    & \multicolumn{1}{l|}{SGPN~\cite{wang2018sgpn}}       & 100\%        & -         & 11.3       & 22.2       & 4.9       & 14.3       & 39.0       \\
			\multicolumn{1}{c|}{}                        & \multicolumn{1}{l|}{3D-SIS~\cite{hou20193d}}     & 100\%        & -         & 18.7       & 35.7       & 16.1      & 38.2       & 55.8       \\
			\multicolumn{1}{c|}{}                        & \multicolumn{1}{l|}{MTML~\cite{lahoud20193d}}       & 100\%        & 20.3      & 40.2       & 55.4       & 28.2      & 54.9       & 73.1       \\
			\multicolumn{1}{c|}{}                        & \multicolumn{1}{l|}{PointGroup~\cite{jiang2020pointgroup}} & 100\%        & 34.8      & 56.9       & 71.3       & 40.7      & 63.6       & 77.8       \\
			\multicolumn{1}{c|}{}                        & \multicolumn{1}{l|}{HAIS~\cite{chen2021hierarchical}}       & 100\%        & 43.5      & 64.1       & 75.6       & 45.7      & 69.9       & 80.3       \\
			\multicolumn{1}{c|}{}                        & \multicolumn{1}{l|}{SSTNet~\cite{liang2021instance}}     & 100\%        & 49.4      & 64.3       & 74.0       & 50.6      & 69.8       & 78.9       \\ 
			\multicolumn{1}{c|}{}                        & \multicolumn{1}{l|}{SoftGroup~\cite{vu2022softgroup}}     & 100\%        & 46.0      & 67.6       & 78.9       & 50.4      & \textbf{76.1}       & \textbf{86.5}       \\
			\multicolumn{1}{c|}{}                        & \multicolumn{1}{l|}{GraphCut~\cite{hui2022graphcut}}     & 100\%        & \textbf{52.2}      & \textbf{69.1}       & \textbf{79.3}       & \textbf{55.2}      & 73.2       & 83.2       \\ \midrule
			\multicolumn{1}{c|}{\multirow{4}{*}{Weakly Sup.}} & \multicolumn{1}{l|}{SPIB~\cite{liao2021point}}       & 0.16\%       & -         & 38.6       & 61.4       & -         & -          & 63.4       \\
			\multicolumn{1}{c|}{}                        & \multicolumn{1}{l|}{SegGroup~\cite{tao2020seggroup}}   & 0.02\%       & 23.4      & 43.4       & 62.9       & 24.6      & 44.5       & 63.7       \\
			\multicolumn{1}{c|}{}                        & \multicolumn{1}{l|}{Baseline}   & 0.02\%       & 21.2      & 39.0       & 61.3       & -         & -          & -          \\
			\multicolumn{1}{c|}{}                        & \multicolumn{1}{l|}{3D-WSIS~(\textbf{ours})}       & 0.02\%       & \textbf{28.1}      & \textbf{47.2}       & \textbf{67.5}       & \textbf{25.1}       & \textbf{47.0}       & \textbf{67.8}       \\ \bottomrule
		\end{tabular}
	}
	\label{tab:ScanNet_val}
\end{table}

\begin{figure}[!t]
	\centering
	\includegraphics[width=0.95\textwidth]{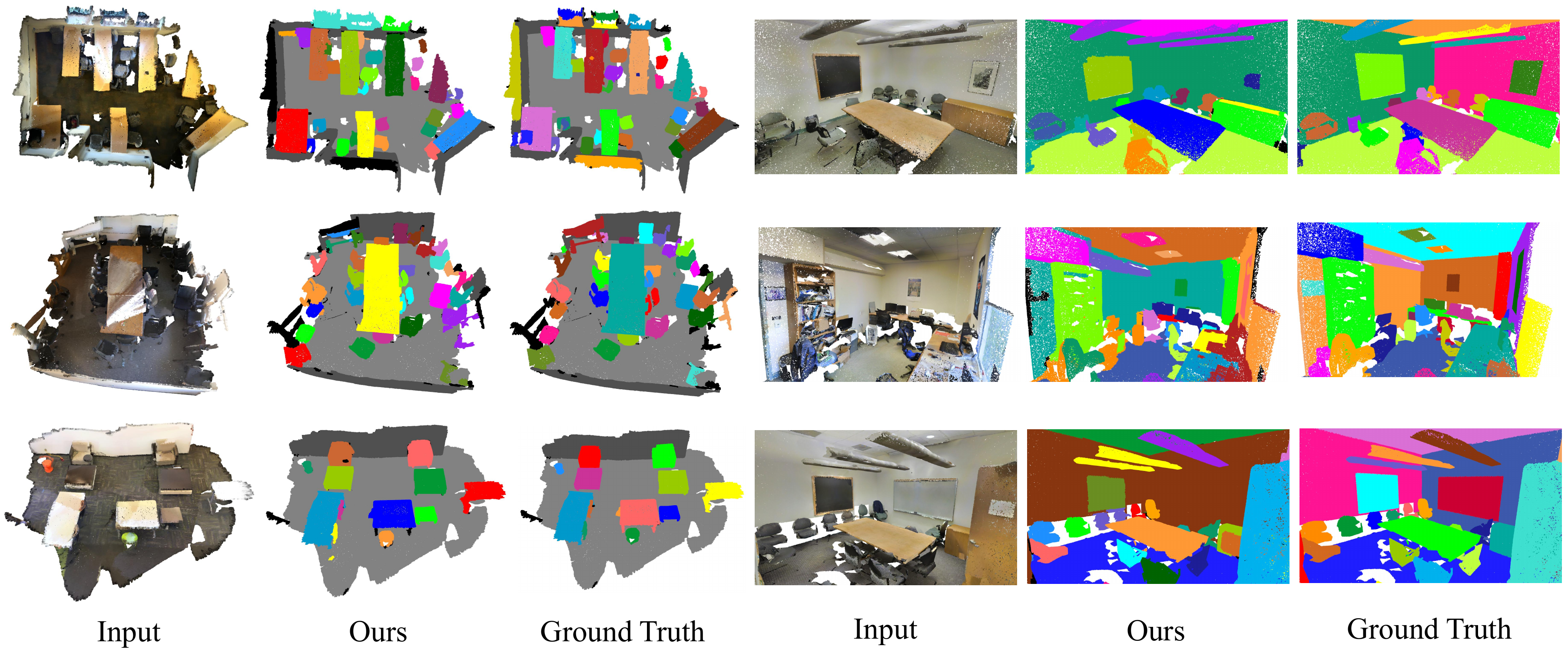}
	\caption{Visualization of the 3D instance segmentation results on the validation of ScanNet-v2 (left) and S3DIS (right). We randomly select colors for different instances.}
	\label{fig:scannet_s3dis_ins}
\end{figure}

\begin{table}[!t]
	
	\centering
	\caption{3D instance segmentation results on the S3DIS dataset. ``Baseline'' means the model trained with the initial annotated labels only.}
	\resizebox{0.95\textwidth}{!}{
		\begin{tabular}{@{}ccc|ccc|cccc@{}}
			\toprule
			\multicolumn{2}{c|}{Method}                                                  &\;Annotation\;   &\;\;AP\;\;  &\;$\text{AP}_{50}$\; &\;$\text{AP}_{25}$\;   & \;mCov & mWCov         & mPrec         & \;mRec\; \\
			\midrule
			\multicolumn{10}{c}{6-fold Cross Validation}                                                                                                       \\ \midrule
			\multicolumn{1}{c|}{\multirow{6}{*}{Fully Sup.}}   & \multicolumn{1}{l|}{SGPN~\cite{wang2018sgpn}}     &100\%  & - & - & -    & 37.9          & 40.8          & 38.2          & 31.2          \\
			\multicolumn{1}{c|}{}                        & \multicolumn{1}{l|}{ASIS~\cite{wang2019associatively}}    &100\%  & - & - & -     & 51.2          & 55.1          & 63.6          & 47.5          \\
			\multicolumn{1}{c|}{}                        & \multicolumn{1}{l|}{PointGroup~\cite{jiang2020pointgroup}} &100\%  & - & 64.0 & -   & -             & -             & 69.6          & 69.2          \\
			\multicolumn{1}{c|}{}                        & \multicolumn{1}{l|}{HAIS~\cite{chen2021hierarchical}}   &100\%   & - & - & -    & 67.0 & 70.4 & 73.2          & 69.4          \\
			\multicolumn{1}{c|}{}                        & \multicolumn{1}{l|}{SSTNet~\cite{liang2021instance}}   &100\%  & 54.1 & 67.8 & -    & -             & -             & 73.5 & 73.4 \\ 
			\multicolumn{1}{c|}{}                        & \multicolumn{1}{l|}{SoftGroup~\cite{vu2022softgroup}}   &100\%  & 54.4 & \textbf{68.9} & -    & 69.3             & 71.7             & \textbf{75.3} & 69.8 \\
			\multicolumn{1}{c|}{}                        & \multicolumn{1}{l|}{GraphCut~\cite{hui2022graphcut}}   &100\%  & \textbf{56.3} & 68.2 & -    & \textbf{72.8}             & \textbf{75.0}             & 74.4 & \textbf{73.7} \\ \midrule
			\multicolumn{1}{c|}{\multirow{2}{*}{Weakly Sup.}} & \multicolumn{1}{l|}{Baseline} &0.02\%  & 19.5 & 30.5 & 42.0  & 41.1          & 42.3          & 13.3           & 37.1          \\
			\multicolumn{1}{c|}{}                        & \multicolumn{1}{l|}{SegGroup} &0.02\%  & 23.1 & 37.6 & 48.5 & 45.5 & 47.6 & 56.7 & 43.3 \\
			\multicolumn{1}{c|}{}                        & \multicolumn{1}{l|}{3D-WSIS~(\textbf{ours})} &0.02\%  & \textbf{26.7} & \textbf{40.4} & \textbf{52.6} & \textbf{48.0} & \textbf{50.5} & \textbf{59.3} & \textbf{46.7} \\ \midrule
			\multicolumn{10}{c}{Area 5}                                                                                                                        \\ \midrule
			\multicolumn{1}{c|}{\multirow{6}{*}{Fully Sup.}}   & \multicolumn{1}{l|}{SGPN~\cite{wang2018sgpn}}  &100\%  & - & - & -     & 32.7          & 35.5          & 36.0          & 28.7          \\
			\multicolumn{1}{c|}{}                        & \multicolumn{1}{l|}{ASIS~\cite{wang2019associatively}}  &100\%  & - & - & -      & 44.6          & 47.8          & 55.3          & 42.4          \\
			\multicolumn{1}{c|}{}                        & \multicolumn{1}{l|}{PointGroup~\cite{jiang2020pointgroup}}  &100\%  & - & 57.8 & -  & -             & -             & 61.9          & 62.1          \\
			\multicolumn{1}{c|}{}                        & \multicolumn{1}{l|}{HAIS~\cite{chen2021hierarchical}}  &100\%   & - & - & -    & 64.3 & 66.0 & 71.1 & 65.0 \\
			\multicolumn{1}{c|}{}                        & \multicolumn{1}{l|}{SSTNet~\cite{liang2021instance}}  &100\%   & 42.7 & 59.3 &  -   & -             & -             & 65.5          & 64.2          \\ 
			\multicolumn{1}{c|}{}                        & \multicolumn{1}{l|}{SoftGroup~\cite{vu2022softgroup}}  &100\%   & 51.6 & 66.1 &  -   & 66.1             & 68.0             & 73.6          & 66.6          \\
			\multicolumn{1}{c|}{}                        & \multicolumn{1}{l|}{GraphCut~\cite{hui2022graphcut}}  &100\%   & \textbf{54.1} & \textbf{66.4} &  -   & \textbf{67.5}             & \textbf{68.7}             & \textbf{74.7}          & \textbf{67.8}          \\ \midrule
			\multicolumn{1}{c|}{\multirow{2}{*}{Weakly Sup.}} & \multicolumn{1}{l|}{Baseline}  &0.02\%  & 18.9 & 26.8 & 37.7   & 36.3   & 37.1          & 11.1           & 28.5          \\
			\multicolumn{1}{c|}{}
			& \multicolumn{1}{l|}{SegGroup}  &0.02\%  & 21.0 & 29.8 & 41.9 & 39.1 & 40.8 & 47.2 & 34.9 \\
			\multicolumn{1}{c|}{}
			& \multicolumn{1}{l|}{3D-WSIS~(\textbf{ours})}  &0.02\%  & \textbf{23.3} & \textbf{33.0} & \textbf{48.3} & \textbf{42.2} & \textbf{44.2} & \textbf{50.8} & \textbf{38.9} \\
			\bottomrule
		\end{tabular}
	}
	\label{tab:S3DIS}
\end{table}

\subsection{Results}

{\bf ScanNet-v2.} Tab.~\ref{tab:ScanNet_val} reports the quantitative results on the ScanNet-v2 validation set and hidden testing set. Compared with the existing semi-/weakly supervised point cloud instance segmentation methods, our approach achieves state-of-the-art performance and improves the $\text{AP}_{25}$ from 62.9\% to 67.5\% with a gain of about 5\% on the ScanNet-v2 validation set. Note that SPIB~\cite{liao2021point} uses the bounding box of each instance as weak labels, which is different from ours. Although we have the same number of annotated instances, clicking one point per instance provides less information than eight corners of the box. Nonetheless, our method can still achieve higher performance than SPIB. 

{\bf S3DIS.} Tab.~\ref{tab:S3DIS} reports the results on Area 5 and 6-fold cross validation of the S3DIS dataset. Compared with the fully supervised 3D instance segmentation methods, our model can still achieve good results, even outperforming the fully supervised methods such as SGPN~\cite{wang2018sgpn}. The quantitative results demonstrate the effectiveness of our method on weakly supervised 3D instance segmentation. %

{\bf Visualization results.} Fig.~\ref{fig:scannet_s3dis_ins} shows the visualization results of instance segmentation on the ScanNet-v2 validation set and S3DIS. It can be found that although some objects of the same class are close to each other, like chairs, the instances are still segmented properly. Since we additionally predict the size of the objects in the network, our method can effectively use the size information of the objects to guide the clustering, thereby segmenting different instances that are close to each other.

\subsection{Ablation Study}
{\bf Effect of pseudo labels.} We conduct experiments on the ScanNet-v2 validation set to verify the effectiveness of our label propagation. The quantitative results are reported in Tab.~\ref{tab:abl_performance}. ``Baseline'' indicates that our method trained with initial annotated labels, without pseudo labels. In the first stage, we report the mean average precision at different iterations, respectively. It can be observed that the performance gradually increases as the number of iterations increases. Furthermore, based on the stage one (dubbed ``Stage 1''), it can be found that the performance is greatly improved after training in the second stage (dubbed ``Stage 2''). Since the number of generated pseudo labels in the first stage is still less than that of fully annotated labels, it is difficult to effectively segment instances. In the second stage, we use the model trained in the first stage to cluster pseudo instances, so we can regard the obtained object volume as the additional supervision to train the network. The quantitative results in the second stage further demonstrate that using the predicted object volume can indeed improve the performance of weakly supervised 3D instance segmentation.

\begin{table}[t]
	\centering
	\caption{The ablation study of different components on the ScanNet-v2 validation set and S3DIS Area 5. ``Baseline'' means the model trained the initial annotated labels only. Note that ``Stage 2'' is performed based on ``Stage 1''.
	}
	\resizebox{0.95\textwidth}{!}{
		\begin{tabular}{@{}cc|ccc|ccccccc@{}}
			\toprule
			\multicolumn{2}{c|}{}                                       & \multicolumn{3}{c|}{ScanNet-v2 Val} & \multicolumn{7}{c}{S3DIS Area 5}                  \\ \midrule
			\multicolumn{2}{c|}{Settings}                               & \;\;AP\;\;        & $\text{AP}_{50}$       & \;$\text{AP}_{25}$\;       & \;\;AP\;\; & $\text{AP}_{50}$ & \;$\text{AP}_{25}$\; & mCov & mWCov & mPrec & mRec \\ \midrule
			\multicolumn{2}{c|}{Baseline}                               & 21.2      & 39.0     & 61.3     &  18.9  &  26.8  &  37.7  &  36.3    &   37.1    &   11.1    &   28.5   \\ \midrule
			\multicolumn{1}{c|}{\multirow{3}{*}{\;Stage 1\;}} & \;Iter. 1\; & 23.4      & 42.2     & 62.8     &  19.9  &  27.7  &  40.2  &  38.5    &   39.4    &   21.7    &  33.1    \\
			\multicolumn{1}{c|}{}                         & \;Iter. 2\; & 24.5      & 43.6     & 64.4     &  20.1  &  28.0  &  40.7  &   39.7   &   40.4    &   22.2    &  33.8    \\
			\multicolumn{1}{c|}{}                         & \;Iter. 3\; & 25.4      & 45.3     & 65.8     &  20.9  &  28.3  &  41.9  &   40.0   &   40.8    &   23.1    &   34.1   \\ \midrule
			\multicolumn{2}{c|}{Stage 2 }                                & \textbf{28.1} & \textbf{47.2} & \textbf{67.5}  &  \textbf{23.3}  &  \textbf{33.0}  &  \textbf{48.3}  &  \textbf{42.2}   &  \textbf{44.2}  &  \textbf{50.8}  & \textbf{38.9}   \\ \bottomrule
		\end{tabular}
	}
	\label{tab:abl_performance}
\end{table}

\begin{table}[t]
	\centering
	\caption{The ablation study results (proportion/accuracy) of pseudo labels at different iterations on the ScanNet-v2 training set. The proportion and accuracy of pseudo labels are computed at the point level.}
	\resizebox{0.95\textwidth}{!}{
		\begin{tabular}{@{}c|ccc@{}}
			\toprule
			\;\;\;\;Stage 1\;\;\;\; & \;\;\;\;Only Random\;\;\;\; & Random+Affinity\;\;  &  Random+Affinity+Semantic  \\ \midrule
			Iter. 1 & 33.7~/~39.9 & 29.1~/~52.7 & 18.2~/~81.9  \\
			Iter. 2 & 47.7~/~38.4 & 35.1~/~71.6 & 30.9~/~82.1  \\
			Iter. 3 & 48.2~/~38.3 & 35.2~/~73.1 & 31.4~/~82.5  \\
			\bottomrule
		\end{tabular}
	}
	\label{tab:abl_label}
\end{table}

{\bf Quality of pseudo labels.} The pseudo labels are generated on the training set to increase supervision during training, so their quality affects network training. To further study the quality of the generated pseudo labels, we count the proportion and accuracy of pseudo labels on the ScanNet-v2 training set during training. The results are listed in Tab.~\ref{tab:abl_label}. When only using random walk (dubbed ``Only Random''), the proportion of pseudo labels is high, but the accuracy is low (39.9\% at ``Iter. 1''). The low-accuracy pseudo labels will affect the training of the network. If we add the extra affinity constraint (dubbed ``Random+Affinity''), we can observe that the proportion of pseudo labels is lower, but the accuracy is greatly improved (52.7\% at ``Iter. 1''). Due to the affinity constraint, the proportion of wrong label propagation is reduced. Therefore, the quality of pseudo labels is improved and high-quality supervision is provided for network training. Furthermore, when we add the semantic constraints (dubbed ``Random+Affinity+Semantic''),  the accuracy of pseudo labels improves from 52.7\% (``Random+Affinity'') to 81.9\% (``Random+Affinity+Semantic''), which shows that the semantic constraint is useful for the weakly supervised 3D instance segmentation task. As constraints are added gradually, the proportion of the generated pseudo labels decreases, while the accuracy increases. 

%

\textbf{Label propagation times.} Different label propagation times influence the quality of pseudo labels. As shown in Tab.~\ref{tab:abl_label}, the proportion and accuracy of pseudo labels at three iterations (dubbed ``Iter. 3'') is comparable to two iterations (dubbed ``Iter. 2''), and performing more iterations consumes more resources, thus we choose three iterations for label propagation.

\begin{figure*}[t]
	\centering
	\includegraphics[width=0.98\textwidth]{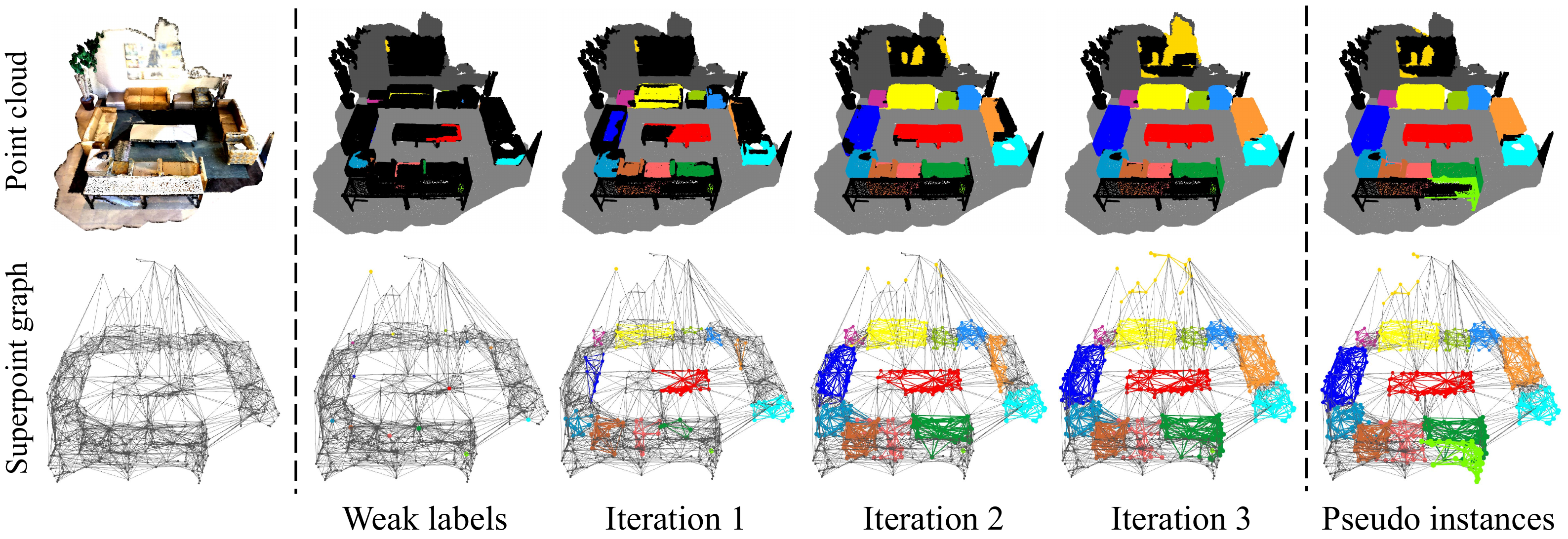}
	\caption{Visualization results of pseudo label generation. Note that we remove the superpoints on the walls and floor for a better view.}
	\label{fig:label_prop}
\end{figure*}

{\bf Visualization of pseudo labels.} Fig.~\ref{fig:label_prop} shows the pseudo labels at different iterations in the first stage on the superpoint graph. We can observe that the initial annotated labels are extremely sparse, and only a few superpoints are annotated in the superpoint graph. In the first stage, as the iteration number of label propagation increases, the labels spread to the surrounding superpoints in the graph gradually. With the constraints of the predicted affinity and semantic, the propagation of labels is restricted to the same object. In the second stage, we use the model trained in the first stage to predict pseudo instances by performing clustering on the superpoint graph. The last column shows the predicted pseudo instances. It can be observed that different instances can be effectively separated. 


\section{Conclusion}
In this paper, we proposed a simple yet effective method for weakly supervised 3D instance segmentation with extremely few labels. To exploit few point-level annotations, we used an unsupervised point cloud oversegmentation method on the point cloud to generate superpoints and construct the superpoint graph. Based on the constructed superpoint graph, we developed an inter-superpoint affinity mining module to adaptively learn inter-superpoint affinity for label propagation via random walk. We further developed a volume-aware instance refinement module to guide the superpoint clustering on the superpoint graph by learning the object volume information. Experiments on the ScanNet-v2 and S3DIS datasets demonstrate that our method achieves state-of-the-art performance on weakly supervised 3D instance segmentation.


\bibliographystyle{splncs04}
\bibliography{ref}

\end{document}